# A Modular Dual-Arm Robotic Cell for Disassembly and Repair of Industrial Control Electronics

Maximilian Ruhe[1], Fabian Harlacher[1], Christian Friedrich[1] and Martin Kipfmüller[1]

[1] Karlsruhe University of Applied Sciences, Karlsruhe, Germany

**Abstract.** Industrial control electronics such as programmable logic controllers, servo drives and operator panels are routinely repaired in plant maintenance, but were never designed for automated disassembly. This paper presents a modular dual-arm robotic cell for repair-oriented disassembly, using two collaborative manipulators, interchangeable tools, red-green-blue-depth (RGB-D) and wrist-level perception, force/torque sensing and a Robot Operating System (ROS) 2-based control with Behavior Tree (BT) execution, teleoperation, digital-twin support and bounded learning-based contact skills. The process is decomposed into sequence planning, symbolic execution with fallbacks, force-limited tool skills, visual condition assessment and demonstration-based adaptation. A CAD-derived device graph encodes the disassembly order, access constraints, tools, feasible removal directions and verification states and converts them into operation objects for the BT and motion layers. Grounded in three representative devices, the cell covers screw removal, damaged-fastener fallback, snap-fit opening, connector release, cooperative printed circuit board (PCB) extraction and condition-based repair decisions. The main contribution is an architecture linking sequence knowledge, perception, verification and force-aware skills through one ROS 2 interface across simulation, teleoperation and real hardware.

**Keywords:** robotic disassembly; repair automation; dual-arm manipulation; sequence planning; behavior trees; digital twin

## 1 Introduction

Circular production requires processes that preserve products, modules and components before they are reduced to material streams. In industrial plant engineering, controllers, drives and operator panels are often removed because of localized defects, maintenance-driven replacement or obsolete configuration, although large parts of the device remain usable. Manual disassembly and repair are therefore part of maintenance, yet these devices arrive with heterogeneous geometries, uncertain wear states, incomplete documentation and no robot-oriented service features. Automated disassembly of such devices differs from automated assembly because the product state is not guaranteed: housings may be deformed, screw heads rounded, polymer clips aged and internal routing altered by previous repairs. Many operations are contact-rich, snap-fits must release without breaking brittle plastic, edge-connector printed circuit boards (PCBs) must come out without bending, and ribbon connectors must unplug within narrow force envelopes. Purely scripted motions are brittle while purely vision-driven policies fail when the decisive signal is force rather than visible geometry. The central thesis is that practical disassembly for industrial electronics requires deterministic structure where it is reliable and adaptive behavior where uncertainty dominates. Structure comes from device-family models, precedence relations, tool selection, controller interfaces, safety envelopes and Behavior Tree (BT) fallbacks; adaptation comes from perception, local pose refinement, teleoperated intervention and learned contact-rich skills.

The paper presents the design and first implementation of a repair-oriented dual-arm cell. Hardware, digital twin and Robot Operating System (ROS) 2 stack share one command and observation interface across simulation, teleoperation and physical execution. The emphasis remains on system architecture; full experimental validation is ongoing. The contributions are: a repair-oriented dual-arm cell for industrial control electronics; a repair toolbox for housings, screws, snap-fits, plug connections, PCBs and inspection; CAD-supported sequence planning as the high-level task model; operation objects carrying target frames, approach directions, force limits and verification criteria; and a shared ROS 2 interface that keeps demonstrations, learned skills and validation episodes comparable.

## 2 Related Work

Reviews of robotic disassembly and Waste Electrical and Electronic Equipment (WEEE) automation identify product variability, uncertain fasteners, contact-rich operations and the laboratory-to-industry gap as dominant open issues [1-3]. Implemented cells often target electric-vehicle batteries, where economic relevance is high and pack or module structure provides clear process stages [4,5], while electronics-focused efforts combine metrology, AI and learned skills for non-destructive disassembly of consumer and office electronics [6]. In contrast, the present work targets condition-aware, repair-oriented disassembly of recurring industrial control electronics, where the goal can shift from component recovery to targeted repair based on detected defects.

Disassembly Sequence Planning (DSP) is well established: classical feature-based methods derive collision-free sequences and removal directions from CAD-based relational models [7,8] and have been extended toward executable manipulation under partial uncertainty [9]. Recent graph-based approaches address interaction uncertainty and online replanning [10]. For repair-oriented industrial electronics, however, the planner must also expose tool access, removal direction, resource usage and verification state to execution.

Behavior Trees structure long-horizon execution by separating local recovery from global task logic [11,12], and skill frameworks such as SkiROS2 add reusable symbolic capabilities with typed planning-execution interfaces [13]. Bimanual teleoperation provides demonstrations for skills that are hard to specify geometrically: ALOHA/ ACT, GELLO and ROS 2-compatible virtual reality (VR) frameworks show how such data can be collected for contact-rich manipulation [14-16]. Contact-rich subtasks - snap-fit release, connector unplugging and PCB extraction - hinge on force response, for which variable impedance gives the control foundation [17]. Diffusion policy and adaptive compliance policy then combine visuomotor policies with learned compliance, enabling bounded adaptation rather than purely scripted contact behavior [18,19]. Perception adds screw detection and PCB-defect inspection [20,21], while sim-to-real manipulation and GPU-parallel simulation support contact-skill development under randomized friction, contact and lighting [22,23]. The gap addressed here is the integration into one repair-oriented loop: device-graph reasoning, force-aware Behavior Tree skills and condition-aware verification share operation objects and common interfaces. Compared with recycling-oriented cells and stand-alone DSP methods, repair preservation is treated here as the organizing constraint, so operations must be geometrically feasible, force-limited and verifiable with respect to the device state. This is why the operation object is positioned as the central interface between sequence knowledge and executable robot behavior, rather than as a mere data container.

## 3 Target Devices and Repair Requirements

The cell is anchored in three representative devices that are repaired in practice but were not designed for robotic service (Fig. 1). The Siemens TP177B operator panel represents layered human-machine interface (HMI) disassembly with front bezel, touch foil, LCD module, gaskets, carrier, rear housing, connectors and main PCB, where display preservation and soft constraints such as adhesion or cable routing are decisive. The Siemens S7-300 programmable logic controller (PLC) represents compact controller electronics with brittle snap-fit housings, side clips, internal PCB structures and plug interfaces. The Lenze EVS-9325 servo inverter represents screw-fastened covers, carrier and shield plates, power and control PCBs, edge-card connectors and power components, combining rigid tool operations with compliant cooperative PCB extraction.

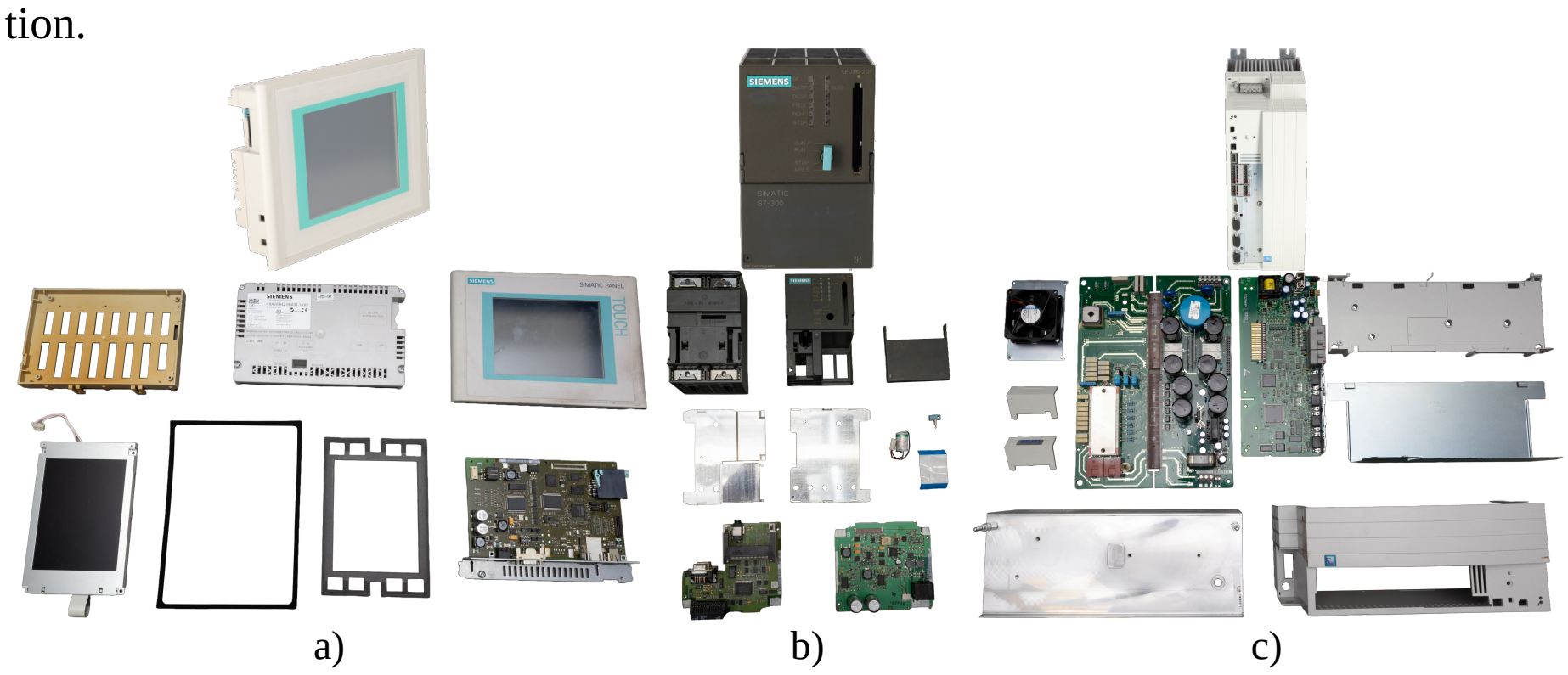

**Fig. 1.** The three representative target devices in assembled (top) and exploded (bottom) views: (a) Siemens TP177B operator panel; (b) Siemens S7-300 CPU315-2 DP PLC; (c) Lenze EVS-9325 servo inverter

These devices lead to five design requirements. R1 is repair-preserving access for housings, boards, displays and connectors. R2 is tool flexibility for screws, clips, connectors, boards and damaged-fastener fallback. R3 is bimanual resource reasoning for single-arm, fixture-assisted and cooperative extraction. R4 is perception-driven verification after each operation. R5 is data continuity across real execution, teleoperation and simulation.

## 4 Hardware Cell and Repair Toolbox

The physical cell consists of two 6-DOF UR15 manipulators with 15 kg payload, 1300 mm reach and wrist-level force/torque sensing, mounted on a common aluminium-profile workbench and operating in a shared workspace (Fig. 2). Bimanuality is treated as a standard resource: one arm may stabilize a housing or board while the second performs a tool operation, or both arms may act synchronously during cooperative extraction. This reduces the need for dedicated mechanical clamping fixtures and provides a reconfigurable, force-monitored alternative for fragile or variable parts. In practice, this is important for aged housings, where over-constraining a plastic shell can cause cracks, and for large PCBs, where asymmetric extraction forces can bend the substrate or damage edge-card connectors.

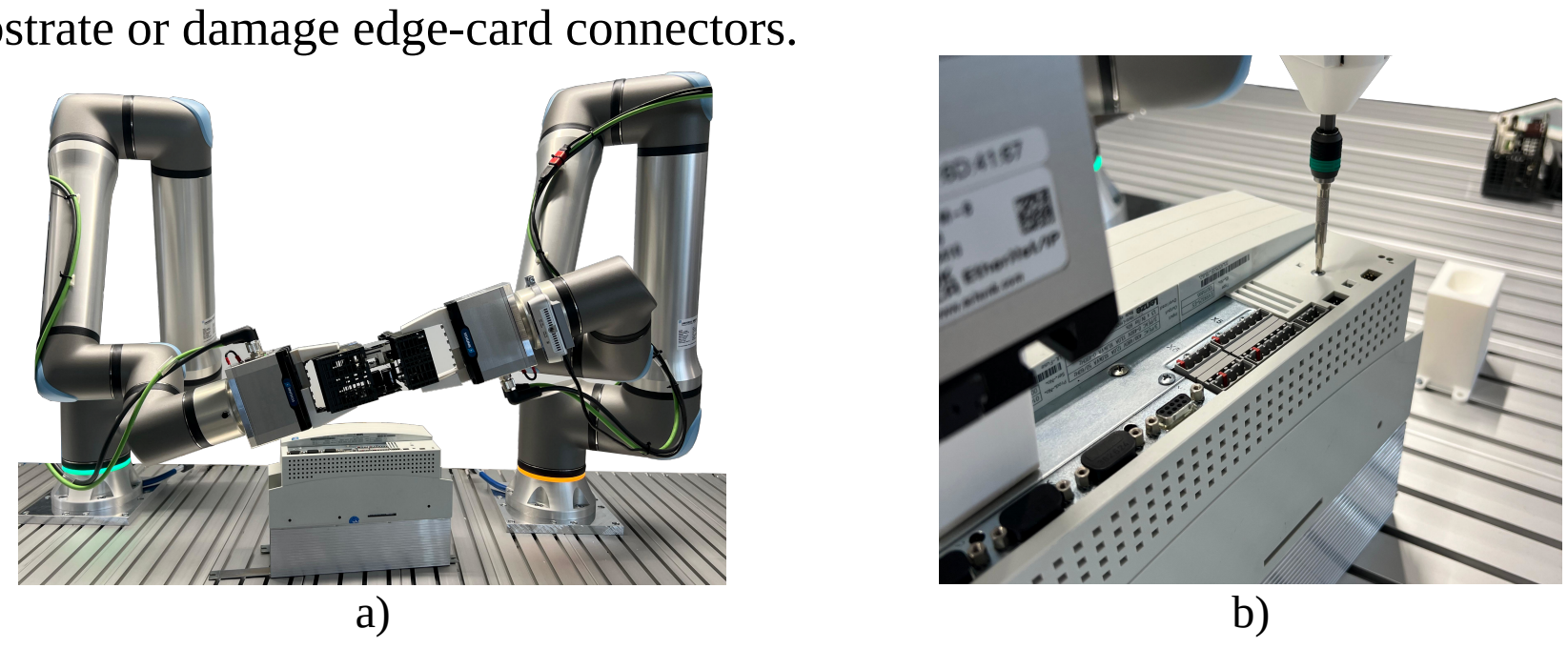

a) b)

**Fig. 2.** The physical cell: (a) two 6-DOF UR15 manipulators with wrist-level force/torque sensing on a shared aluminium-profile workbench, configured for cooperative handling; (b) close-up of a single-arm screw-removal operation on a device stabilized by the second arm.

The proposed framework builds on a reusable repair toolbox that abstracts recurring disassembly and repair operations into transferable manipulation skills. Instead of designing device-specific procedures, the system decomposes repair processes into reusable operation primitives that can be combined across heterogeneous products and varying device geometries. The current implementation employs a compact set of generic primitives that serve as the basis for higher-level operations such as fastening removal, connector handling, housing opening, component extraction, and inspection. By grounding manipulation in reusable operation objects rather than product-specific workflows, the same skill set can be applied across different industrial devices while allowing task-specific parameterization through CAD information, perception, and process context. This modular abstraction facilitates scalability to new product families and enables future extension toward additional manipulation capabilities without re-designing the overall planning framework.

# 5 Software Architecture

The software architecture (Fig. 3) is centered on ROS 2. Sensors, robot drivers, controllers, tools, perception modules, teleoperation interfaces and simulation endpoints share one command and observation path, avoiding divergent simulation, teleoperation and hardware stacks.

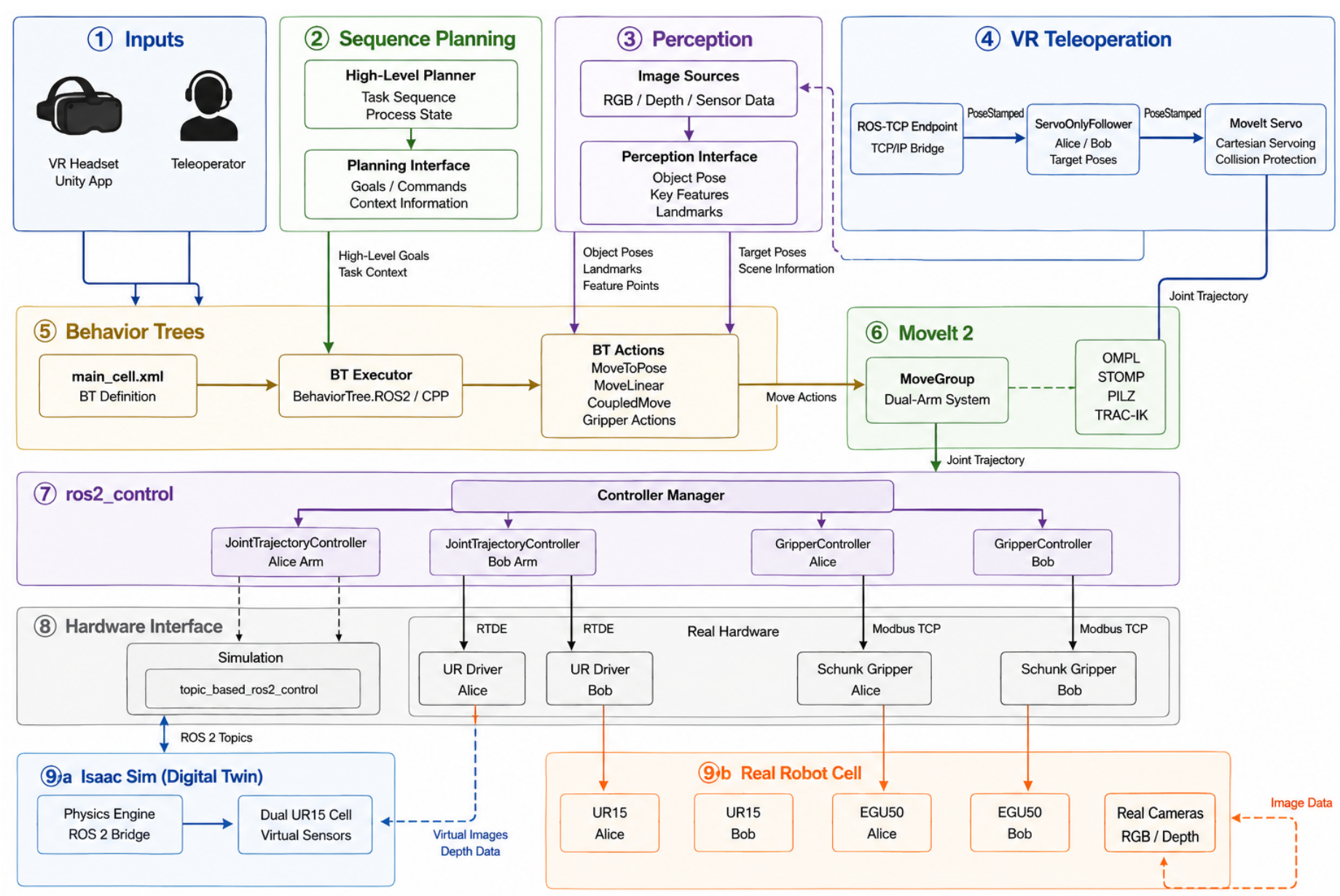

**Fig. 3.** ROS 2-based system architecture. A task layer (sequence planning + Behavior Tree, BT, execution), fed by perception, the CAD device graph, VR teleoperation and learned skills, issues motion goals to MoveIt 2; ros2_control then drives, through one controller manager, *either* the Isaac Sim digital twin (via the ROS 2 bridge) *or* the real cell — UR15 arms over RTDE and Schunk grippers over Modbus/TCP. Command and observation messages are identical across simulation and hardware. RGB-D: colour-plus-depth; RTDE: real-time data exchange.

At the conceptual level perception, the CAD-derived device graph, teleoperation, digital-twin services and learned skills feed a task layer that combines sequence planning and Behavior Tree execution before actions are routed through ROS 2 control. The same operation representation is used for simulation, teleoperation, autonomous execution and replay.

Concretely, the stack forms one path from decision to actuator command. Planner and Behavior Tree executor run as ROS 2 nodes; a motion leaf requests a trajectory from MoveIt 2 (single-arm or coupled dual-arm), executed through ros2_control. Its controller manager is the single switch point between simulation and hardware: the same controllers drive either the Isaac Sim twin over the ROS 2 bridge or the real UR15 arms (RTDE) and Schunk grippers (Modbus/TCP). Perception and wrist force/torque flow back along the same path, feeding the skills and the force-limit fallbacks and updating the operation object; each leaf thus closes a local loop of plan, move, sense and verify and returns an outcome that may trigger replanning. Because twin and cell exchange identical messages, a subtree validated in simulation runs unchanged on hardware, with only contact-sensitive parameters recalibrated.

## 6 CAD-Supported Sequence Planning and Operation Objects

Sequence planning is based on a CAD-derived relational assembly model in which components, geometric constraints and process dependencies are represented as a graph. Building on prior work [9, 24], the relative disassembly spaces between components are transformed into global removal spaces through visibility-based reasoning; the underlying disassembly-space computation scales efficiently for structured CAD-derived assemblies. In the present architecture, this model is not used as a static offline list: online scene information updates accessibility, component state and verification labels, and deviations such as inaccessible fasteners, unexpected screw damage or a blocked connector trigger replanning.

The planner reasons not only about geometric feasibility but also about execution constraints: tool access, preferred approach directions, force limits, verification requirements and whether an action is single-arm, fixture-assisted or cooperative. CAD, service information and manual annotations provide the transparent initial model, while perception and execution outcomes provide runtime corrections. For a screw, the graph stores the expected pose, head type, access direction and cover dependency; for a snap-fit, it stores the clip node, lever direction and force envelope; for a PCB, it stores grasp regions, connector line and permitted extraction axis. This makes the approach suitable for recurring industrial device families whose nominal structure is known but whose individual units vary in wear, condition and accessibility.

To bridge symbolic planning and executable robot behavior, each task is converted into an operation object (Table 1), executed by the Behavior Tree layer, which returns symbolic outcomes such as success, blocked, damaged, excessive force or failure.

**Table 1.** Operation-object fields passed from planning to execution.

| Field | Meaning | Example |
|---|---|---|
| Target element | part, fastener, connector or inspection node | side-panel screw; PLC clip; Lenze power PCB |
| Tool class | required end effector or process tool | screwdriver; lever; gripper; milling spindle |
| Manipulation Primitive | Robot action to perform (e.g. UNSCREW, LEVER, PULL) | Unscrew a screw |
| Pose and approach | target frame, preferred access direction and local approach constraint | wrist approach normal to screw head; lever tip aligned with clip slot |
| Force and safety limits | admissible wrench, speed and workspace envelope | maximum lever force; differential PCB force |
| Verification condition | observable postcondition for graph update | screw removed; clip released; board extracted |

For example, a snap-fit opening operation carries the clip node as target element, the polymer lever as tool, the insertion direction and an admissible force range. Verification is a wrist-camera check that the clip released without fracture, and the fallback stops the skill and requests re-approach when the force limit is exceeded. A cooperative PCB extraction instead carries the board grasp regions, the edge-card connector axis as removal direction, a differential-force limit and a dual-arm resource class. A scheduling tick computes the reachable frontier of the device graph, attaches perception labels and confidence values, allocates the required tool and resource class, and dispatches executable operations. After execution, the returned outcome updates the graph and may trigger a fallback or a new planning tick. Sequence planning thus becomes the high-level memory of the process rather than a static list of robot poses.

## 7 Execution, Perception and Adaptive Skills

Execution is delegated to a Behavior Tree layer. Each operation object is realized by a compact subtree containing precondition checks, tool selection, motion planning, skill execution, postcondition verification and fallback handling. Fallbacks are encoded structurally: a bit slip or damaged screw head can dispatch a milling fallback; excessive force during levering stops the skill and requests re-approach; and a physically blocked graph node returns a blocked outcome to the planner. This is essential for legacy devices, where damaged fasteners, brittle housings and changed internal routing are expected events rather than exceptions. Scripted skills, impedance-controlled skills, teleoperated actions and learned policies can all be invoked as long as they obey the same safety contract and return the same outcome vocabulary.

Perception is organized in two stages. Scene-level RGB-D perception localizes the device, candidate screws, connectors, clips and exposed boards. After a coarse approach, wrist-level perception refines the active manipulation target by classifying screw-head type, damage, connector family or local landmarks. Here the repair branch denotes the alternative process path the cell enters when inspection reveals a localized, repairable defect: rather than continuing to fully disassemble the device for component recovery, the cell isolates and exposes only the affected part for targeted repair (for example, replacing a damaged capacitor) while leaving healthy modules assembled. The repair branch adds PCB-level inspection for visible defects such as damaged capacitors, charred regions, lifted traces, missing components or solder anomalies. These outputs are not used as isolated detections; they update operation objects and verification states. A positive defect detection can switch the process from disassembly for recovery to targeted repair while preserving healthy modules. Representative local operations and inspection targets are shown in Fig. 4.

Simulation is a central development environment. Omniverse Isaac Sim mirrors the robot cell, device geometry and sensor viewpoints and connects through the same ROS 2 control interface as the physical cell. The twin supports reachability checks, collision testing, viewpoint planning, Behavior Tree debugging and synthetic data generation. Contact-rich skills are pre-trained in the twin under randomized conditions before they are calibrated on hardware, and the VR front end doubles as an operator-training view.

High-level control and Behavior Tree code are reused across simulation and hardware, while contact-sensitive parameters still require real-world calibration. Isaac Lab supports GPU-parallelized training with randomized friction, contact stiffness, lighting and fastener condition [23].
Teleoperation supports runtime intervention and offline demonstration capture. VR is suitable for general bimanual actions and operator training, while a GELLO-style leader-arm path is useful for contact-rich subtasks requiring felt response, such as lever insertion, connector release or PCB extraction. Demonstrations record target poses, forces, perception context and outcomes for imitation learning or policy refinement. Learned contact-rich skills are invoked only where geometry alone is insufficient. The reference skill for contact-rich subtasks is the Adaptive Compliance Policy (ACP) [19]. Impedance control lets the robot behave like a programmable spring whose stiffness determines how strongly it follows a target pose. ACP augments a diffusion-policy backbone with visual, proprioceptive and force/torque observations and predicts a reference pose, a virtual target pose and a stiffness value. Their relation determines the compliant direction, allowing high stiffness for accurate tracking and lower stiffness where contact forces must be accommodated.

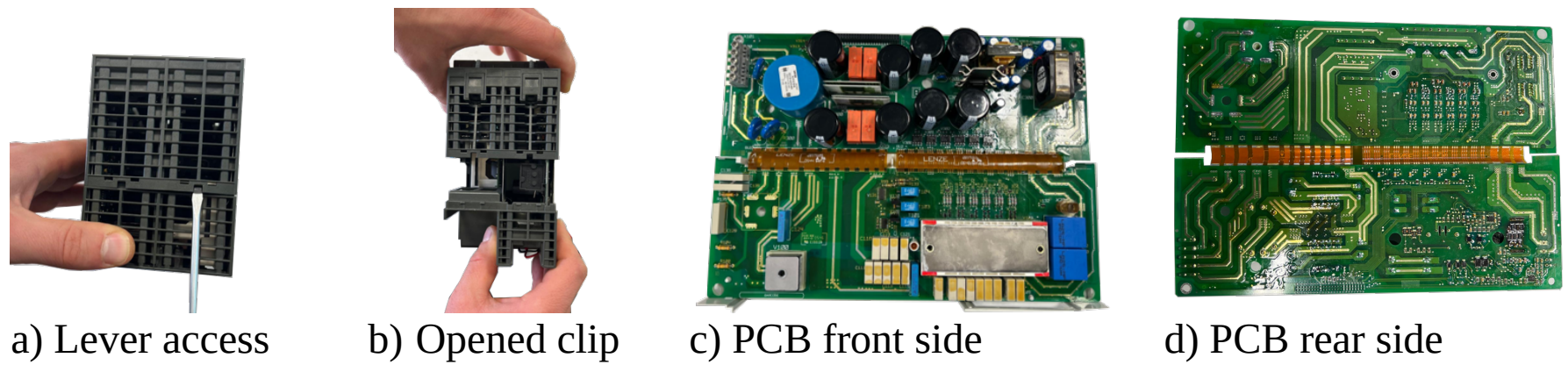

a) Lever access b) Opened clip c) PCB front side d) PCB rear side

**Fig. 4.** Repair-relevant local operations and inspection targets: (a) lever access to a snap-fit; (b) opened-clip state after force-limited levering; (c, d) front and rear side of an extracted PCB used for defect inspection (e.g. damaged capacitors, charred regions, lifted traces).

## 8 Integration, Demonstration and Evaluation Plan

The three devices instantiate complementary stress cases for the same planning-execution contract. The S7-300 PLC tests force-limited snap-fit opening, where clip order, lever direction and force response determine success. The Lenze servo drive tests the transition from rigid screw operations to cooperative PCB extraction along edge-card connector axes, where differential-force monitoring guards against asymmetric loading. The TP177B operator panel tests annotation-driven sequencing for layered display modules with soft constraints such as adhesion, gaskets and cable routing. In all cases, the device graph selects the next reachable operation, the Behavior Tree executes it with explicit postcondition checks and perception updates whether the process should continue, inspect, recover or change to a repair branch.
The current implementation comprises the dual-arm hardware, Isaac Sim twin, ROS 2 control stack, CAD-derived device graphs and first tool-based operations. Demonstrations include screw detection and removal, cover opening, lever-based snap-fit access with learned compliance, teleoperated intervention and demonstration capture (Figs. 2 and 4). One recurring experience is that Behavior Tree logic and perception viewpoints prepared in the twin transfer to the cell unchanged, while contact-sensitive force and stiffness parameters require on-hardware recalibration. The present status demonstrates feasibility of integration, while quantitative evaluation is ongoing.
Planned evaluation separates subtask from complete-cycle performance. Metrics include screw-removal success, damaged-fastener detection, snap-fit opening without visible fracture, connector-removal success, cooperative PCB-extraction force asymmetry, perception precision and recall, PCB-defect detection, tool-change time, cycle time and recovery branches. Baselines include scripted fixed-stiffness actions, teleoperated assistance and learned contact-rich skills; sequence-planning metrics include graph-completion efficiency, replanning events and adaptation effort to new variants. Limitations are recurring device-family dependence, initial graph creation and sim-to-real calibration for aged plastic, friction and damaged fasteners.

## 9 Conclusion

This paper presented a modular dual-arm robotic cell for repair-oriented disassembly of industrial control electronics. Its central contribution is the system-level integration around operation objects, linking device-graph reasoning, tool-based and force-aware Behavior Tree execution, perception-driven verification, teleoperation and digital-twin development in one repair-oriented loop across simulation and physical hardware.
The architecture addresses a practically relevant class of legacy industrial devices that are routinely repaired in maintenance workflows but were not engineered for automated disassembly. From a circular production perspective, it targets a high-value segment of the repair chain by preserving embedded material and manufacturing effort and by creating a scalable pathway for integrating electronics repair into circular factory workflows.

## 10 Acknowledgments

This research is being conducted as part of the project “NewEra”. The project is funded by the Federal Ministry of Research, Technology and Space (BMFTR) within the funding measure “Digital GreenTech – Environmental Technology Meets Robotics”, which is part of the BMFTR strategy “Research for Sustainable Development (FONA)”